\documentclass[11pt]{article}

\usepackage[round]{natbib}
\usepackage{graphicx}
\usepackage{fullpage}

\begin{document}

\title{A Study of the Learning Progress in Neural Architecture Search Techniques}

\author{
Prabhant Singh\thanks{prabhantsingh@gmail.com} \and
Tobias Jacobs\thanks{tobias.jacobs@neclab.eu} \and
S\'{e}bastien Nicolas\thanks{sebastien.nicolas@neclab.eu} \and
Mischa Schmidt\thanks{mischa.schmidt@neclab.eu}
}

\date{\textit{NEC Laboratories Europe GmbH, Heidelberg, Germany}\\June 6, 2019}


\maketitle

\begin{abstract}
In neural architecture search, the structure of the neural network to best model a given dataset is determined by an automated search process. Efficient Neural Architecture Search (ENAS), proposed by~\cite{pham2018efficient}, has recently received considerable attention due to its ability to find excellent architectures within a comparably short search time. 
In this work, which is motivated by the quest to further improve the learning speed of architecture search, we evaluate the learning progress of the controller which generates the architectures in ENAS. We measure the progress by comparing the architectures generated by it at different controller training epochs, where architectures are evaluated after having re-trained them from scratch.
As a surprising result, we find that the learning curves are completely flat, i.e., there is no observable progress of the controller in terms of the performance of its generated architectures. This observation is consistent across the CIFAR-10 and CIFAR-100 datasets and two different search spaces. We conclude that the high quality of the models generated by ENAS is a result of the search space design rather than the controller training, and our results indicate that one-shot architecture design is an efficient alternative to architecture search by ENAS.
\end{abstract}


\section{Introduction}

Neural architecture search deals with the task of identifying the best neural network architecture for a given prediction or classification problem. Treating the network architecture as a black box that can be evaluated by first training and then testing it, a systematic try-and-error process has been demonstrated to discover architectures that outperform the best human-designed ones for a number of standard benchmark datasets~\citep{zoph2017neural}.

As each single black-box evaluation of an architecture involves training a deep neural network until convergence, it is attractive to use a faster heuristic to predict the performance of candidate architectures. The ability  to still find a near-optimal solution depends on the correlation between the predicted and the actual performance of the candidate. A remarkably simple and attractive performance estimation method has been recently proposed by~\cite{pham2018efficient}. In their method, called \emph{Efficient Neural Architecture Search (ENAS)}, the whole search space of possible architectures utilizes a common set of shared weights. These weights are trained simultaneously with a controller learning to generate architectures.

As the ENAS search process has been demonstrated to find very good architectures within very reasonable computational constraints, we are using it as the basis for our studies, which have the goal to further reduce the search time. As a starting point, we evaluate the architectures generated by ENAS over the course of the search process, in order to receive indication about when the \emph{exact} architecture performance converges. In this work the exact performance of an architecture shall be defined as its test accuracy after having trained its weights for a sufficient number of epochs with the full training set. 

As a very surprising result, we did not find any measurable improvement in terms of the exact performance over the course of the search. In other words, the architectures generated by the ENAS controller at the beginning of the search performed just as well as the architectures generated after hundreds of controller training epochs. Our observation has been consistently confirmed across runs of the experiment for the CIFAR-10 and CIFAR-100  dataset \citep{Krizhevsky2009learning} with the two distinct search spaces described by \cite{pham2018efficient}. While our results are inconsistent with the ablation studies presented in \citep{pham2018efficient}, they are in line with a recent series of findings about the search performance of ENAS \citep{Li2019random, Adam2019understanding}.

Our results support the hypothesis that excellent neural architectures can be obtained by a one-shot approach, and, with an appropriate design space and probability distribution, one-shot methods are a serious alternative to ENAS-like search methods. We remark that already ENAS could be seen as a one-shot method, because during its search it never evaluates the exact performance of any candidate architecture before making its final decision. Our work demonstrates that skipping the ENAS search process, which takes several GPU-hours, does not diminish the quality of the result. 

A brief overview of related work is given in the subsequent section. In Section~\ref{sec:background} we outline the ENAS method which is re-evaluated in this work, using the experimental setup described in Section~\ref{sec:experiments}. The results of our experiments are also presented and discussed in Section~\ref{sec:experiments}, and we conclude this work in Section~\ref{sec:conclusion}.

\section{Related work}
\label{sec:related_work}


The idea of using machine learning methods to determine neural network architectures has been popularized by the seminal work of~\cite{zoph2017neural}, who have demonstrated that \emph{Neural Architecture Search} (NAS) is able to find networks that outperform the best human-designed neural architectures. The most obvious drawback of the original NAS method is its resource requirement, as the learning process involves generating a large number of deep neural networks, whose performance is evaluated by fully training them with the given training dataset and evaluating them with the test set, which takes several hours or even days per candidate architecture.

There is a large body of follow-up work studying attempts to speed up the search. Much attention has been gained by~\cite{pham2018efficient}, who have proposed the Efficient Neural Architecture Search method based on the idea to design the search space such that each candidate network is a subgraph of a joint network. The weights of the joint network are trained while an architecture generator (controller) is trained to select the best sub-network for the given dataset. Using this approach, the authors were able to improve the time complexity from thousands of GPU days to less than a single GPU day for the CIFAR-10 dataset while reaching a performance similar to the original NAS results. 

Numerous alternative methods to ENAS have been presented as well.
\cite{Suganuma2018genetic} are among the authors who have provided evolutionary methods for finding neural architectures.
The search method described by \cite{Liu2018progressive} starts with evaluating simpler architectures and gradually makes them more complex, using a surrogate function to predict the performance of candidate architectures. 
\cite{Liu2008darts} employ a continuous relaxation of the network selection problem, which enables the search procedure to use gradients to guide the search. 
A continuous search space is also employed by \cite{Luo2018neural}. Here a model is trained to estimate the performance of an architecture based on a representation in a continuous space, providing a gradient based on which the architecture can be improved. A comprehensive survey of neural architecture search methods has been published by \cite{Elsken2019neural}.

One-shot approaches for finding neural architectures have been described in several recent works. \cite{Brock2018one} have proposed to have the weights of candidate neural architectures generated by a hypernetwork that is trained only once. A simpler mechanism inspired by the ENAS approach has been presented by \cite{Bender2018understanding}. Here the joint network containing all weights is first trained, and then the validation accuracy using these weights is used to predict the performance of candidate architectures. Instead of using a reinforcement learning based controller, the candidate architectures are generated by random search.
The latter two approaches are applying search techniques which enumerate many possible architectures, but they can be considered one-shot methods in the sense that only one architecture is fully trained. In this sense, also ENAS can be considered a one-shot method.
In contrast, our results support the hypothesis a single architecture generated at random from an appropriate search space - thus applying one-shot learning in a stricter sense - has competitive performance to architectures resulting from search processes.

Our findings are consistent with the very recent results of \cite{Adam2019understanding}, who have analyzed the behavior of the ENAS controller during search and found that its hidden state does not encode any properties of the generated architecture, providing an explanation of the observation made by \cite{Li2019random} that ENAS does not perform better than simple random architecture search. 
Our work is complementing that line of research by providing an experimental study of the learning progress of ENAS, demonstrating that good performance is already achieved before any search has taken place.
This is also in consistent with recent results of \cite{Xie2019Exploring}, who have experimented with neural network generators for image recognition and have found that, without search, the randomly generated networks achieve state-of-the-art performance.

\section{Background}
\label{sec:background}

In this work we study the learning progress of the ENAS procedure as presented by~\cite{pham2018efficient}, systematically evaluating early and late architectures that are generated during the search process. 

Candidate architectures, so-called \emph{child models}, are generated by ENAS using a \emph{controller}, which is modeled as an LSTM with 400 hidden units. The architecture search space is designed such that all child models are subgraphs of a \emph{joint network}; each child model is defined as a subgraph containing a subset of the joint network's nodes and edges. Importantly, the trainable weights are shared across the child models. Whenever a child model has been generated by the controller, the weights contained by it are updated using the gradient from a single training minibatch. The performance of the child model on the validation set is then used as a feedback signal to improve the controller.

For the CIFAR-10 dataset, \cite{pham2018efficient} experiment with two distinct search spaces. In the \emph{macro search space} the controller samples architectures layer by layer, deciding about the layer type (3x3 or 5x5 convolution, 3x3 or 5x5 depthwise-separable convolution, average pooling or max pooling) and the subset of lower layers to form skip connections with. The set of available operations  as well as the number of layers is defined by hand.

In the \emph{micro search space}, the network macro-architecture in terms of interconnected convolutional and reduction \emph{cells} is defined by hand, while the controller decides about the computations that happen inside each cell. Inside a convolutional cell, there are a fixed number of nodes, where each node can be connected with two previous nodes chosen by the controller and performs an operation among identity, 3x3 or 5x5 separable convolution, average pooling, or max pooling. Reduction cells are generated using the same search space, but with the nodes applying their operations with a stride of 2. The cell output is obtained by concatenating the output of all nodes that have not been selected as inputs to other nodes.

\cite{pham2018efficient} also design a dedicated search space for designing recurrent neural networks for language processing tasks. We do not include experiments with this search space in our experiments, but note that similar findings can be expected in light of recent studies of the ENAS search process in \citep{Li2019random, Adam2019understanding}.

\section{Experiments}
\label{sec:experiments}
For our experiments we are using the original implementation of ENAS as provided by \cite{pham2018efficient} at \texttt{https://github.com/melodyguan/enas/}. This repository provides an implementation ready to be used to search for a CNN for the CIFAR-10 dataset as well as an implementation of ENAS search for an RNN architecture. To search for a CNN architecture for CIFAR-100, we only modified the input data pipeline. The existing implementation also included no support for realizing average pooling and max pooling operations in the final training of architectures generated using the macro search space. We have added this feature in our version, which is available at \texttt{https://github.com/prabhant/ENAS-cifar100}.
The experiments were executed using a Nvidia 1080Ti processor.


\begin{figure}[h]
\begin{center}
	\includegraphics[width=0.5\textwidth]{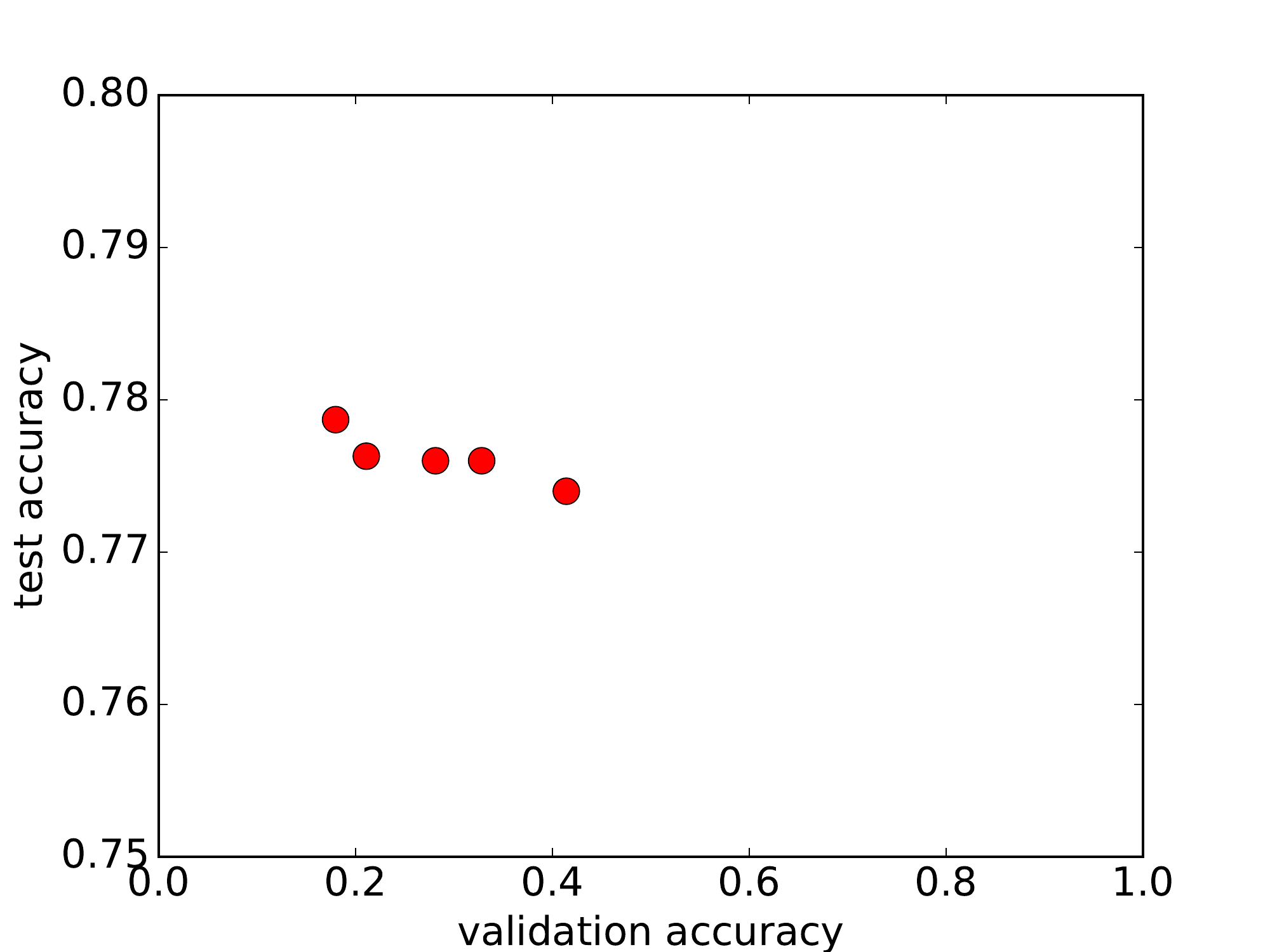}
	\caption{
	\label{fig:scatterplot}
	Scatter plot showing validation accuracy using the shared weights (x-axis) and test accuracy of the architecture after re-training the architecture for 260 epochs. Each point represents a neural architecture generated by the ENAS controller after having been trained for 155 epochs using the macro search space.
	}
\end{center}
\end{figure}

\begin{table}[h]
	\centering	
	\caption{
	\label{tab:cifar}
	Performance of architectures generated by ENAS for CIFAR-10 and CIFAR-100. For the macro search space, the controller was trained for 310 epochs. For the micro search space, it was trained for 150 epochs. Before testing, all architectures were re-trained for 310 epochs. Table cells with multiple numbers correspond to multiple separate runs of the same experiment.
	}
	\begin{tabular}{|c|c||c|c|}
		\hline
		data & search space & initial accuracy & final accuracy \\
		\hline
		CIFAR-10 & macro & {96.69\% / 95.80\% / 95.71\%} & {95.38\% / 95.81\% / 95.70\%} \\
		CIFAR-10 & micro & {94.56\%} & {94.59\%} \\
		CIFAR-100 & macro & {77.12\% / 80.75\% / 80.55} & {80.07\% / 80.39\% / 80.47\%} \\
		CIFAR-100 & micro & {79.59\% / 77.69\%} & {80.20\% / 80.50 \%} \\
		\hline
	\end{tabular}
\end{table}

In Figure~\ref{fig:scatterplot} we take a snapshot of the macro search space controller for CIFAR-100 at training epoch 155 (i.e. after half of the training has been done) and compare the validation accuracies (used for further controller training) of some generated architectures with the test accuracies of the same architectures after having re-trained them. We find that in this sample there is no positive correlation between these two metrics, and thus we cannot expect that the validation accuracy with shared weights represents a useful training feedback for controller improvement.

Table~\ref{tab:cifar} shows the test accuracy after having re-trained architectures generated by the ENAS controller before and after the architecture search procedure. In each experiment run, we have trained the controller using the ENAS procedure. In each row we compare the pair of architectures achieving the best validation accuracy at the first and after the last epoch of controller training, but we perform the comparison in terms of the test accuracy after re-training. Note that the validation accuracy in the first epoch is computed with the randomly initialized shared weights, while the validation accuracy after the last epoch is computed using the shared weights that have been trained during the search process. The way of selecting the architecture after the last epoch corresponds to the final architecture selection method proposed by~\cite{pham2018efficient}.
Most of the experiments were done for the CIFAR-100 dataset, which is more challenging than CIFAR-10 and thus should have a higher potential to show benefits of architecture search.
However, the numbers in the table show that the improvement is below 1\% in the majority of tested configurations. Thus, the training process, which takes several GPU-hours, can often be skipped without sacrificing any accuracy.

\section{Conclusion}
\label{sec:conclusion}
We have evaluated the architecture improvement achieved during the search progress of ENAS, and we found that high-quality architectures are already found before any training of controller and shared weights has started. 
%
Future work will include a reproduction of our results in more experiment runs, using further datasets and including recurrent neural network design. Furthermore, a systematic study on the prediction quality of the validation accuracy (when using tentative weights like in ENAS) will shed more light on the effectiveness of this speedup-technique in general. Finally, the methodology applied in this work can be utilized to evaluate the learning progress of other neural architecture search methods as well.

\bibliographystyle{plainnat}
\bibliography{enas_lc}

\begin{thebibliography}{13}
\providecommand{\natexlab}[1]{#1}
\providecommand{\url}[1]{\texttt{#1}}
\expandafter\ifx\csname urlstyle\endcsname\relax
  \providecommand{\doi}[1]{doi: #1}\else
  \providecommand{\doi}{doi: \begingroup \urlstyle{rm}\Url}\fi

\bibitem[Adam and Lorraine(2019)]{Adam2019understanding}
George Adam and Jonathan Lorraine.
\newblock Understanding neural architecture search techniques.
\newblock \emph{CoRR}, abs/1904.00438, 2019.

\bibitem[Bender et~al.(2018)Bender, Kindermans, Zoph, Vasudevan, and
  Le]{Bender2018understanding}
Gabriel Bender, Pieter{-}Jan Kindermans, Barret Zoph, Vijay Vasudevan, and
  Quoc~V. Le.
\newblock Understanding and simplifying one-shot architecture search.
\newblock In \emph{Proceedings of the 35th International Conference on Machine
  Learning, {ICML} 2018, Stockholmsm{\"{a}}ssan, Stockholm, Sweden, July 10-15,
  2018}, pages 549--558, 2018.

\bibitem[Brock et~al.(2018)Brock, Lim, Ritchie, and Weston]{Brock2018one}
Andrew Brock, Theodore Lim, James~M. Ritchie, and Nick Weston.
\newblock {SMASH:} one-shot model architecture search through hypernetworks.
\newblock In \emph{6th International Conference on Learning Representations,
  {ICLR} 2018, Vancouver, BC, Canada, April 30 - May 3, 2018, Conference Track
  Proceedings}, 2018.

\bibitem[Elsken et~al.(2019)Elsken, Metzen, and Hutter]{Elsken2019neural}
Thomas Elsken, Jan~Hendrik Metzen, and Frank Hutter.
\newblock Neural architecture search: {A} survey.
\newblock \emph{Journal of Machine Learning Research}, 20:\penalty0
  55:1--55:21, 2019.

\bibitem[Krizhevsky and Hinton(2009)]{Krizhevsky2009learning}
Alex Krizhevsky and Geoffrey Hinton.
\newblock Learning multiple layers of features from tiny images.
\newblock Technical report, 2009.

\bibitem[Li and Talwalkar(2019)]{Li2019random}
Liam Li and Ameet Talwalkar.
\newblock Random search and reproducibility for neural architecture search.
\newblock \emph{CoRR}, abs/1902.07638, 2019.

\bibitem[Liu et~al.(2018{\natexlab{a}})Liu, Zoph, Neumann, Shlens, Hua, Li,
  Fei{-}Fei, Yuille, Huang, and Murphy]{Liu2018progressive}
Chenxi Liu, Barret Zoph, Maxim Neumann, Jonathon Shlens, Wei Hua, Li{-}Jia Li,
  Li~Fei{-}Fei, Alan~L. Yuille, Jonathan Huang, and Kevin Murphy.
\newblock Progressive neural architecture search.
\newblock In \emph{Computer Vision - {ECCV} 2018 - 15th European Conference,
  Munich, Germany, September 8-14, 2018, Proceedings, Part {I}}, pages 19--35,
  2018{\natexlab{a}}.

\bibitem[Liu et~al.(2018{\natexlab{b}})Liu, Simonyan, and Yang]{Liu2008darts}
Hanxiao Liu, Karen Simonyan, and Yiming Yang.
\newblock {DARTS:} differentiable architecture search.
\newblock \emph{CoRR}, abs/1806.09055, 2018{\natexlab{b}}.

\bibitem[Luo et~al.(2018)Luo, Tian, Qin, Chen, and Liu]{Luo2018neural}
Renqian Luo, Fei Tian, Tao Qin, Enhong Chen, and Tie{-}Yan Liu.
\newblock Neural architecture optimization.
\newblock In \emph{Advances in Neural Information Processing Systems 31: Annual
  Conference on Neural Information Processing Systems 2018, NeurIPS 2018, 3-8
  December 2018, Montr{\'{e}}al, Canada.}, pages 7827--7838, 2018.

\bibitem[Pham et~al.(2018)Pham, Guan, Zoph, Le, and Dean]{pham2018efficient}
Hieu Pham, Melody Guan, Barret Zoph, Quoc Le, and Jeff Dean.
\newblock Efficient neural architecture search via parameter sharing.
\newblock In \emph{International Conference on Machine Learning}, pages
  4092--4101, 2018.

\bibitem[Suganuma et~al.(2018)Suganuma, Shirakawa, and
  Nagao]{Suganuma2018genetic}
Masanori Suganuma, Shinichi Shirakawa, and Tomoharu Nagao.
\newblock A genetic programming approach to designing convolutional neural
  network architectures.
\newblock In \emph{Proceedings of the Twenty-Seventh International Joint
  Conference on Artificial Intelligence, {IJCAI} 2018, July 13-19, 2018,
  Stockholm, Sweden.}, pages 5369--5373, 2018.

\bibitem[Xie et~al.()Xie, Kirillov, Girshick, and He]{Xie2019Exploring}
Saining Xie, Alexander Kirillov, Ross~B. Girshick, and Kaiming He.
\newblock Exploring randomly wired neural networks for image recognition.
\newblock \emph{CoRR}, abs/1904.01569.

\bibitem[Zoph and Le(2017)]{zoph2017neural}
Barret Zoph and Quoc~V. Le.
\newblock Neural architecture search with reinforcement learning.
\newblock In \emph{5th International Conference on Learning Representations,
  {ICLR} 2017, Toulon, France, April 24-26, 2017, Conference Track
  Proceedings}, 2017.

\end{thebibliography}

\end{document}